\documentclass[runningheads]{llncs}
\usepackage[utf8]{inputenc}
\usepackage[english]{babel}

\usepackage[hyphens]{url}

\usepackage{longtable,booktabs}

\usepackage{graphicx}
\usepackage{xcolor,colortbl}
\usepackage{blindtext}
\usepackage{amsmath}
\usepackage{caption}
\usepackage{multirow}
\usepackage[normalem]{ulem}
\usepackage[misc]{ifsym}

\renewcommand{\vec}{\mathbf}

\begin{document}

\title{Importance of Textlines in Historical Document Classification}

\author{Martin Kišš (\Letter) \orcidID{0000-0001-6853-0508} \and
Jan Kohút\orcidID{0000-0003-0774-8903} \and \\
Karel Beneš\orcidID{0000-0002-0805-1860} \and
Michal Hradiš\orcidID{0000-0002-6364-129X}}

\authorrunning{M. Kišš et al.}

\institute{Brno University of Technology, Czech Republic \\
\email{\{ikiss,ikohut,ibenes,hradis\}@fit.vutbr.cz}}

\maketitle

\begin{abstract}
This paper describes a system prepared at Brno University of Technology for ICDAR 2021 Competition on Historical Document Classification, experiments leading to its design, and the main findings. 
The solved tasks include script and font classification, document origin localization, and dating.
We combined patch-level and line-level approaches, where the line-level system utilizes an existing, publicly available page layout analysis engine.
In both systems, neural networks provide local predictions which are combined into page-level decisions, and the results of both systems are fused using linear or log-linear interpolation. 
We propose loss functions suitable for weakly supervised classification problem where multiple possible labels are provided, and we propose loss functions suitable for interval regression in the dating task.
The line-level system significantly improves results in script and font classification and in the dating task.
The full system achieved 98.48\,\%, 88.84\,\%, and 79.69\,\% accuracy in the font, script, and location classification tasks respectively.
In the dating task, our system achieved a mean absolute error of 21.91 years.
Our system achieved the best results in all tasks and became the overall winner of the competition.

\end{abstract}

\begin{figure}[!ht]
    \centering
    \includegraphics[width=0.76\textwidth]{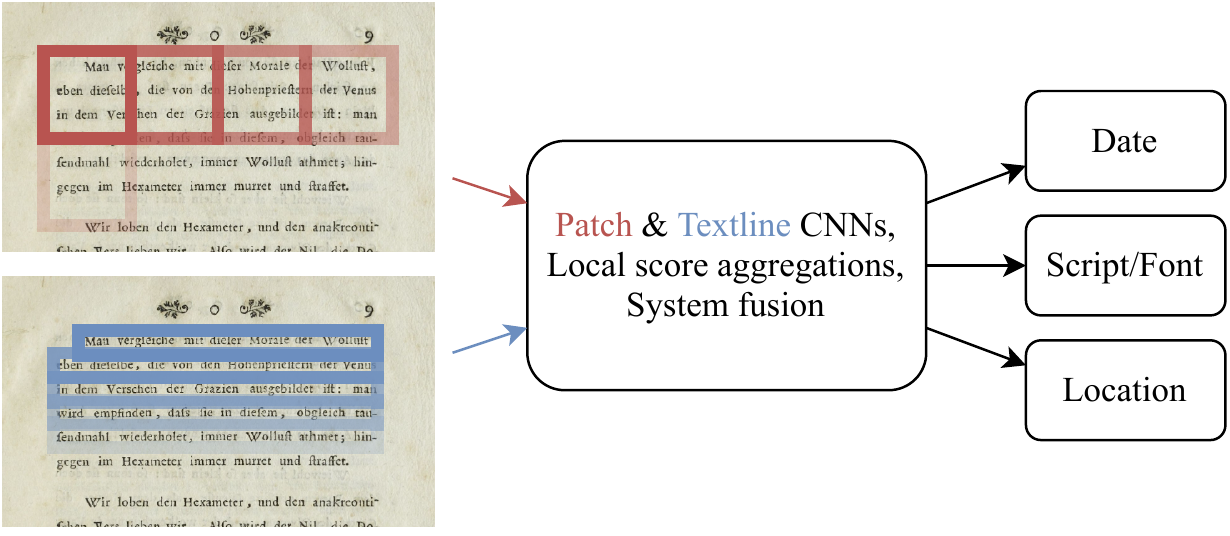}
    \caption{Proposed historical document classification system.    }\label{fig:system_abstract}
\end{figure}

\section{Introduction}

Visual document classification is a fundamental task when working with historical documents.
If one wants to study documents originating from a given location, a given time period or written in a particular font or script, it is first necessary to identify such documents.

This paper describes our approach to the ICDAR 2021 Competition on Historical Document Classification~\cite{seuret_icdar_2021} (HDC).
The challenge targets three different tasks:
(1) Document localization is a simple classification task, where a document needs to be associated with one of the thirteen possible locations of origin.
(2) The goal of font and script classification is to identify typefaces present on a document. The training data does not contain any information about location of individual typefaces and, correspondingly, the output is evaluated only at the page level.
(3) In the dating task, the goal is to estimate the year of production of a given document page. The ground truth annotations are in the form of intervals (not-sooner-than, not-later-than).

Our approach combines two neural networks, where one operates on  square patches of the original documents and the other processes cropped and height-normalized textlines.
The local scores are aggregated into page-level output by simple ad-hoc rules.
For a given task, the page-level outputs of both networks are combined into a final prediction by a trained fusion.

While the models for the localization task are trained using the standard cross entropy, we propose two loss functions to deal with the weakly supervised script and font classification.
As the labels for the dating task are intervals, we train the models using our proposed interval Huber loss function.

In this paper we show that:
\begin{itemize}
    \item the proposed loss functions are effective for weakly supervised classification and interval regression,
    \item the textline processing significantly outperforms patch level processing in the dating and the script and font classification tasks,
    \item the fusion of both models increases the accuracy of the overall system confirming their complementarity.
\end{itemize}

\section{Related work}
Most recent approaches to visual classification of historical documents are based on the use of convolutional neural networks~\cite{cheikhrouhou_multi-task_2021,christlein_deep_2019,cloppet_icdar2017_2017,cloppet_icfhr2016_2016,tensmeyer_convolutional_2017}.
Specifically, these networks are mostly based on the architecture of the ResNet or VGG network~\cite{cheikhrouhou_multi-task_2021,christlein_deep_2019,tensmeyer_convolutional_2017} and their task is to classify individual patches or textlines obtained from the original document.
The individual local outputs often need to be aggregated to obtain a single output for a given page, which is usually done by averaging~\cite{cloppet_icdar2017_2017,cloppet_icfhr2016_2016,tensmeyer_convolutional_2017}.
During training of the network, the input images are often augmented using affine transformations and/or by adding noise.

Christlein et al.~\cite{christlein_deep_2019} proposed a Deep Generalized Max Pooling layer to replace the Global Average Pooling layer at the end of the ResNet architecture, which resulted in accuracy improvement on the CLaMM'16~\cite{cloppet_icfhr2016_2016} and CLaMM'17~\cite{cloppet_icdar2017_2017} competition datasets.

Cheikhrouhou et al. \cite{cheikhrouhou_multi-task_2021} used multi-task learning to train a textline model for keyword spotting and script identification.
In this approach, the model is a sequence of convolutional, recurrent, and fully connected layers.
The output of the last fully connected layer in the main branch is processed by a keyword spotting (KWS) decoder.
Local features, obtained from the convolutional part of the network, and global features, obtained from the recurrent part of the network, are combined in the auxiliary branch using a Compact Bilinear Pooling (CBP) and further processed by fully connected layers to classify the script of the input image.
The identified script is also used as a secondary input to the KWS decoder to further improve its accuracy.
In contrast to this approach, we primarily focus on font and script classification, we compare the textline approach with the patch approach, and we train and evaluate the systems on historical documents.

The most successful approaches in previous historical document classification competitions~\cite{cloppet_icdar2017_2017,cloppet_icfhr2016_2016} use the ResNet architecture operating on patches, using data augmentations comprising of scaling and adding noise.
Apart from CNN-based approaches, some successful systems follow more traditional approach by extracting SIFT descriptors and aggregating them into i-vectors~\cite{cloppet_icfhr2016_2016} or GMM-supervectors~\cite{cloppet_icdar2017_2017}.

\section{Datasets}
For training, validation and testing, we used a dataset published within the ICDAR 2021 Competition on Historical Document Classification~\cite{seuret_icdar_2021}.
The competition consists of three page-level tasks: 1) font group and script type classification, 2) dating, 3) and place of origin estimation.
Overall, the dataset consists of about 70k unique pages.
For most of the pages, only one of font, location, script, or date is annotated; for pages from CLaMM'17~\cite{cloppet_icdar2017_2017} both script and date labels are provided.
Sizes of datasets for the individual tasks are summarized in Table~\ref{tab:datasets}.

\begin{table}[t]
    \centering
    \caption{%
        Number of pages with specific task annotations in the training dataset from ICDAR 2021 Competition on Historical Document Classification.
    }\label{tab:datasets}
    \begin{tabular}{lrrr}
        \toprule
                    Task &
                    Training &
                    Validation &
                    Testing \\
        \midrule
            Font     & 35\,382 &    239 & 5\,506 \\
            Script   &  7\,594 &    419 & 1\,256 \\
            Location &  5\,397 &     65 &    325 \\
            Date     & 10\,294 & 1\,000 & 2\,516 \\
        \bottomrule
    \end{tabular}
\end{table}

As the script, font and dating datasets are not partitioned into training\,--\,valida\-tion subsets, we created our own splits\footnote{The splits are publicly available at \url{https://pero.fit.vutbr.cz/hdc_dataset}}.
For script and font datasets, we created a validation set with uniform class distribution. 
This decision was motivated by our expectation that the test set may contain significantly different class ditribution (however, this is not the case).
For the dating dataset, we randomly selected 1000 pages as the validation set.

\begin{figure}[t]
    \centering
    \includegraphics[width=0.75\textwidth]{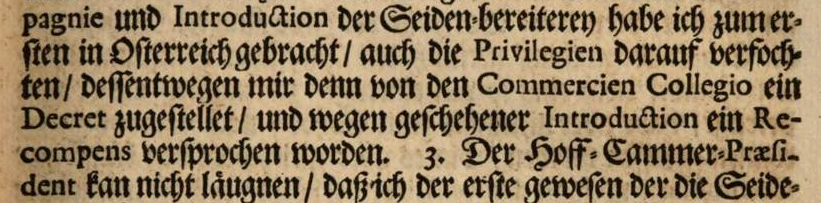}
    \caption{
        An example of a page with mixed fonts.
    }\label{fig:mixed-fonts}
\end{figure}

While there is only one correct label for each page in the location classification, some pages from the script and font classification tasks contain multiple scripts/fonts and are thus assigned multiple labels in the dataset.
This occurs for example in texts written using Fraktur, where some words are written using Antiqua as depicted in Figure~\ref{fig:mixed-fonts}.
Pages in the dating task are annotated with intervals $\langle{}a;b\rangle$ representing the estimated range of years when the document was created.

\section{Document classification systems}

\begin{figure}[t]
    \centering
    \includegraphics[width=\textwidth]{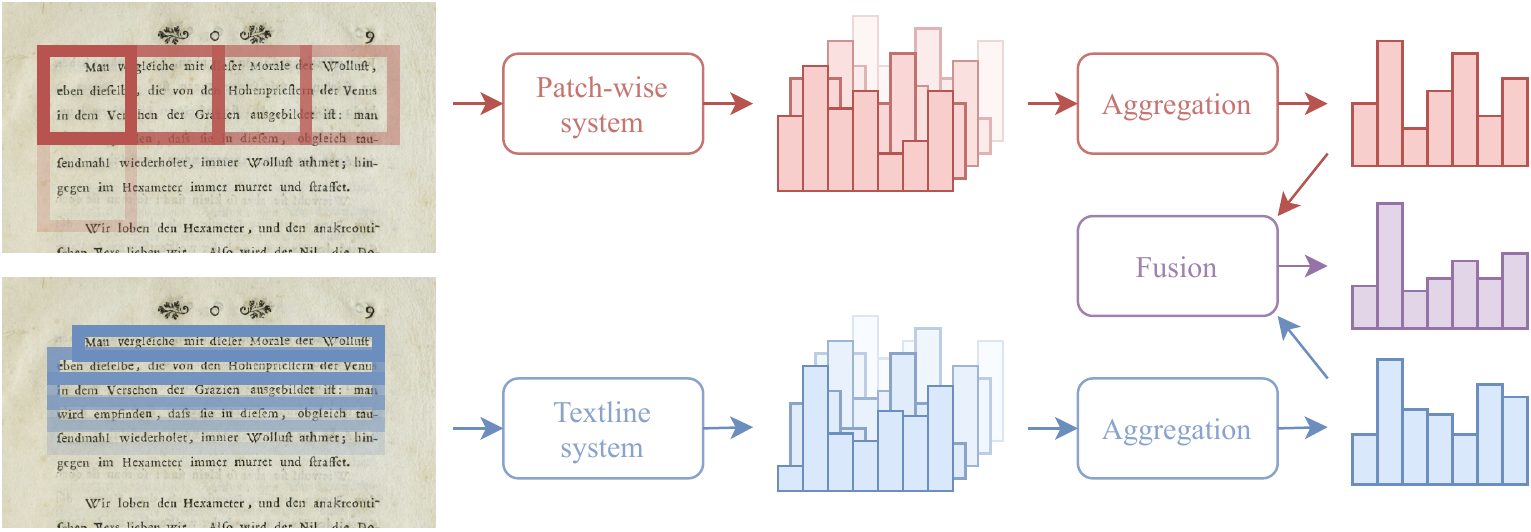}
    \caption{
        Schema of the document classification system. An input page is processed by both the patch system (red) and the textline system (blue). The patch and texline outputs are aggregated into two respective page-level results which are finally fused into a single decision (violet).
    }\label{fig:system}
\end{figure}

Our document classification systems are based on convolutional neural networks processing \emph{cutouts} of the original page images.
One system operates on fixed size rectangular image patches while the second system operates on height-normalized crops of automatically detected textlines.
The type of the systems' output depends on the target task.
For the dating task, the output is a single floating point value which is then transformed to represent the estimated year.
For the rest of the tasks, the output is a vector of floating point values representing a categorical distribution over all possible classes of the given task.
As a the neural networks process each image cutout independently, we aggregate all individual outputs to compute the final page-level output.
During experiments we have tested several patch system approaches and local score aggregation methods for textline system and we describe them together with the neural network architectures in detail in Section \ref{sec:patch-system} and Section \ref{sec:textline-system}, respectively.

\begin{figure}[t]
    \includegraphics[height=0.55cm]{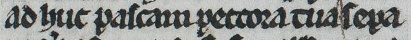}
    \hfill
    \includegraphics[height=0.55cm]{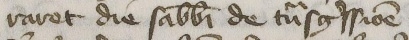}
    \\
    \includegraphics[height=0.55cm]{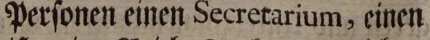}
    \hfill
    \includegraphics[height=0.55cm]{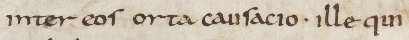}
    \\
    \includegraphics[height=0.55cm]{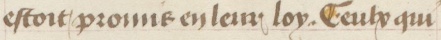}
    \hfill
    \includegraphics[height=0.55cm]{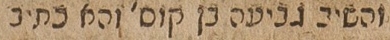}
    \\
    \includegraphics[height=0.55cm]{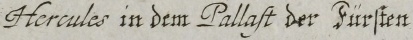}
    \hfill
    \includegraphics[height=0.55cm]{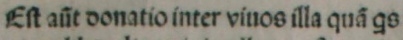}
    
    \caption{Examples of detected and cropped textlines from the dataset.
    }
    \label{fig:detected-lines}
\end{figure}

As both systems utilize information about text placement in a page, we used a detector based on ParseNet architecture~\cite{kodym_page_2021} to detect text regions and individual textlines.
The detector was trained on various documents mainly based on the PERO layout dataset~\cite{kodym_page_2021} and it detected more than 3M textlines in the provided datasets.
Examples of the detected textlines are depicted in Figure~\ref{fig:detected-lines}.

\subsection{Loss functions}
\label{sec:losses}
Except for the localization task, where standard cross entropy loss function is used, the other tasks presented in the challenge don't provide a clear objective to optimize during training.
The difficulty posed in the script and font classification tasks arises from the fact that a single page can be labeled with multiple target classes, e.g.\ because there are multiple scripts present on it.
We propose two loss functions which effectively allow the network to choose one of the page-level labels as relevant for a given cutout. 

Both weakly supervised loss functions are based on cross entropy with respect to each of the page-level labels.
Loss $L_{hard}$ effectively selects the most probable page-level label (based on network output) and ignores the other labels. 
Loss $L_{soft}$ considers all page-level labels weighted by the network output probability. 

If we denote the network output for a single image cutout $x$ as $f(x)$, the predicted probability of class $i$ as $f(x)_i$, and the set of relevant page-level classes as $\mathcal{T}$, the two loss functions are defined as:

\begin{eqnarray}\label{eq:l_hard}
    L_{\mathit{hard}} = \min\limits_{i \in \mathcal{T}} \left[ -\log(f(x)_i) \right]
\end{eqnarray}

\begin{eqnarray}\label{eq:l_soft}
    L_{\mathit{soft}} = \sum\limits_{i \in \mathcal{T}} -\log(f(x)_i) \cdot f(x)_i.
\end{eqnarray}

\begin{figure}[t]
    \centering
    \includegraphics[width=0.66\textwidth]{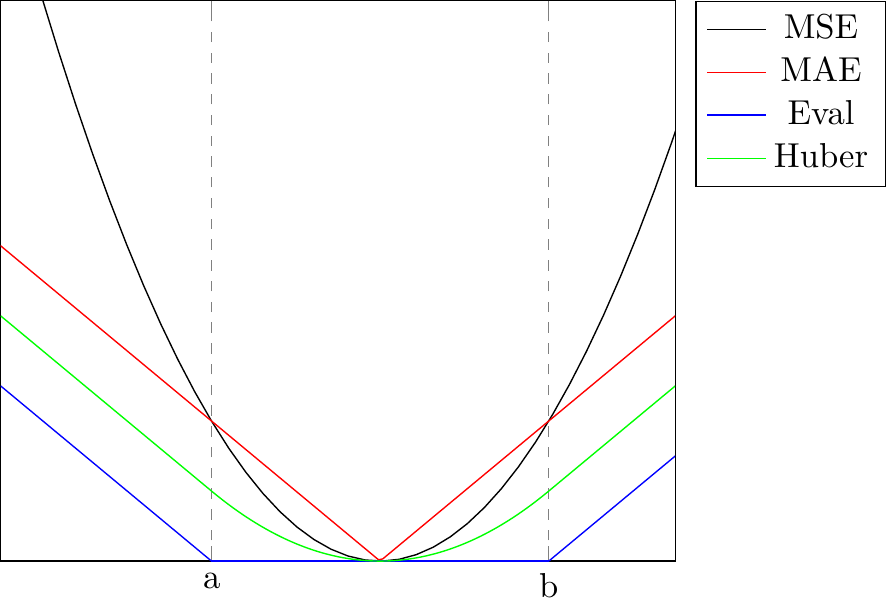}
    \caption{
        Different losses used for date regression.
        \emph{MSE} and \emph{MAE} are the corresponding norm of the distance from the midpoint of the target interval $\langle{}a;b\rangle$.
        \emph{Eval} is the evaluation metric, i.e.\ absolute error from the target interval.
        Finally, \emph{Huber} is the proposed interval Huber loss, which pulls the output of the system towards the midpoint, but the learning signal gradually weakens inside the interval.
    }\label{fig:date-losses}
\end{figure}

In the dating task, the difficulty arises from the target not being a single point in time (year), but rather an interval $\langle{}a;b\rangle$.
The evaluation metric defined in HDC competition  --- mean absolute distance from the interval --- is differentiable and could directly be used as a loss function.
However, it remains open if there really should not be any learning signal provided within the target interval and how do objectives based directly on the interval compare with the more common regression objective functions that drive the output of a system to a single point.
Therefore, we compare the evaluation metric with mean absolute and mean squared error from the midpoint of the target interval that force the output of the network deep inside the label interval.
Additionally, we introduce a interval Huber loss:

\begin{eqnarray}\label{eq:date-loss}
    L_{\mathit{Huber}}(y, a, b)= 
        \begin{cases}
            a-y+\dfrac{r}{2},              & \text{if } y\leq a\\[8pt]
            y-b+\dfrac{r}{2},              & \text{if } y\geq b\\[8pt]
            \dfrac{(y-m)^2}{2r}, & \text{otherwise}
        \end{cases},
\end{eqnarray}
where $y$ is the output of the system, $m = (a+b)/2$ is the midpoint of the target interval, and $r = m -a$ is its radius.
As shown in Figure~\ref{fig:date-losses}, $L_{\mathit{Huber}}$ is linear outside of the ground truth interval as is the evaluation metric, yet it provides a weak learning signal inside the target interval.

\subsection{Patch system}
\label{sec:patch-system}

The patch-level system relies on a ResNeXt-50 neural network~\cite{xie_aggregated_2017} operating on $224\times 224$ patches.
The classification of a page is done by aggregating results of four different image scales, while each scale is divided into a grid of non-overlapping patches.

Patches of each scale are aggregated separately.
We consider either all patches in a whole page approach (P), or only text region patches in a text region approach (R).
Only the ten most confident patches are averaged to give the result for an image scale.
We select the most confident scale as the final page output.

We also combined the whole page approach and the text regions approach, whole page + text regions approach (P+R).
In this case, the final page aggregation is the average of all eight scale aggregations. 
The motivation is that the information inside the text regions should be more valuable for the font and the script tasks but still there can be some useful information outside of these text regions.

\subsection{Textline system}
\label{sec:textline-system}

The textline system operates on arbitrarily long textline images with normalized height of 40 pixels.
The neural network of the textline system uses the convolutional part of VGG16~\cite{simonyan_very_2015} pretrained on ImageNet. 
The covolutional features are aggregated using global average pooling, and processed by three linear layers interleaved with dropout layers and LeakyReLU activation functions.
As each page contains a variable number of textlines, we propose several methods to aggregate these into a single page-level output.

\subsubsection{Classification tasks aggregations.}
We propose three methods to obtain aggregated result for the classification tasks.
The \emph{mean} method calculates a mean vector from the outputs obtained on individual textlines and the final label is identified as the class with the highest probability.
Rest of the methods represent each textline $i$ by its most probable predicted class $c_i$ and its probability $p_{i_c}$.
The \emph{count} method selects most frequently predicted textline class.
The \emph{probs} method selects the class with highest cumulative probability by accumulating the predicted probabilities of the winning textline classes.

\subsubsection{Dating task aggregations.} 
We propose two methods to obtain page-level predictions for the dating task.
The first method computes the \emph{mean} of the results obtained on individual textlines and the second selects the \emph{median} value.

\subsection{System fusion}
To benefit from the different nature of the two systems, we combine them in two fusions.
In both cases, the fusion operates on the page level, i.e.\ it is agnostic to the local score aggregation method.

The first one is a simple linear fusion of the outputs:
\begin{equation}
    \vec{y}_{\mathrm{lin}} = \alpha \vec{y}_{1} + (1 - \alpha) \vec{y}_{2},
\end{equation}
where $\vec{y}_1$ and $\vec{y}_2$ is the full output the patch and textline systems, i.e.\ two vectors of probabilities of all classes relevant for the task.
This approach naturally extends to the date regression, where $\vec{y}_{1,2}$ are scalar date estimates.
For each of the tasks, we tune the interpolation coefficient $\alpha$ independently to on the relevant validation set.

For the classification tasks, we trained a multiclass logistic regression on the outputs of the individual systems:
\begin{equation}
    \vec{y}_{\mathrm{log}} = \mathrm{softmax} ( \vec{W} [ \vec{y}_{1} ; \vec{y}_{2} ] + \vec{b} )
\end{equation}
Here, $[ \vec{y}_{1} ; \vec{y}_{2} ]$ is a concatentation of the class probabilities coming from the invidual systems and $\vec{W}$ and $\vec{b}$ are trained parameters of the fusion.
In order to avoid overfitting, we apply a L2 regularization towards aggressive averaging, i.e.\ towards $\vec{W} = \mathbf{1}$ and $\vec{b} = \mathbf{0}$.
We determine the strength of this regularization by a~10-fold cross-validation on all the available data.
Finally, we train the fusion on all the data.

\section{Experiments}
We conducted experiments with the patch and the textline system on the dataset provided for the HDC competition.
Specifically, we trained both systems using both proposed weakly supervised loss functions $L_{hard}$ and $L_{soft}$ for the script and the font tasks.
For the dating task, we also trained both systems with all four regression loss functions mentioned in Section~\ref{sec:losses}.
For all tasks, we evaluated the proposed patch system approaches and proposed local score aggregation methods for textline system.

\subsection{Experimental setup}

\subsubsection{Patch system training}
The patches were randomly cropped out of the pages and augmented with a small random linear transformation, color jitter and gaussian blur. 
For each task, a system had been trained until it converged, approximately 100k iterations.
We used Adam as the optimization method, the learning rate was progressively enlarged for the first 5k iterations until it reached $3\times 10^{-4}$, the dropout was set to 25\,\%.
Every 20k iterations, the learning rate and dropout were reduced until they hit values of $3.5\times 10^{-5}$ and 0\,\%, respectively.

\subsubsection{Textline-level system training}
Before the training, we removed textlines shorter than 128 pixels to avoid training on false-positive textline detections.
During the training, we used data augmentation comprising of defocusing, adding gaussian noise, and color shifting.
To mitigate the imbalance of the provided datasets we weighted contribution of training samples of each class in our loss functions by inverse frequency of samples of the class.
Optimization of the model was performed using Adam optimizer with learning rate of $1\times 10^{-4}$ for 100k iterations with batch size of 32 and dropout rate 10~\%.

\subsection{Results}

\subsubsection{Dating task} 
\begin{figure}[t]
    \centering
    \includegraphics[width=1\textwidth]{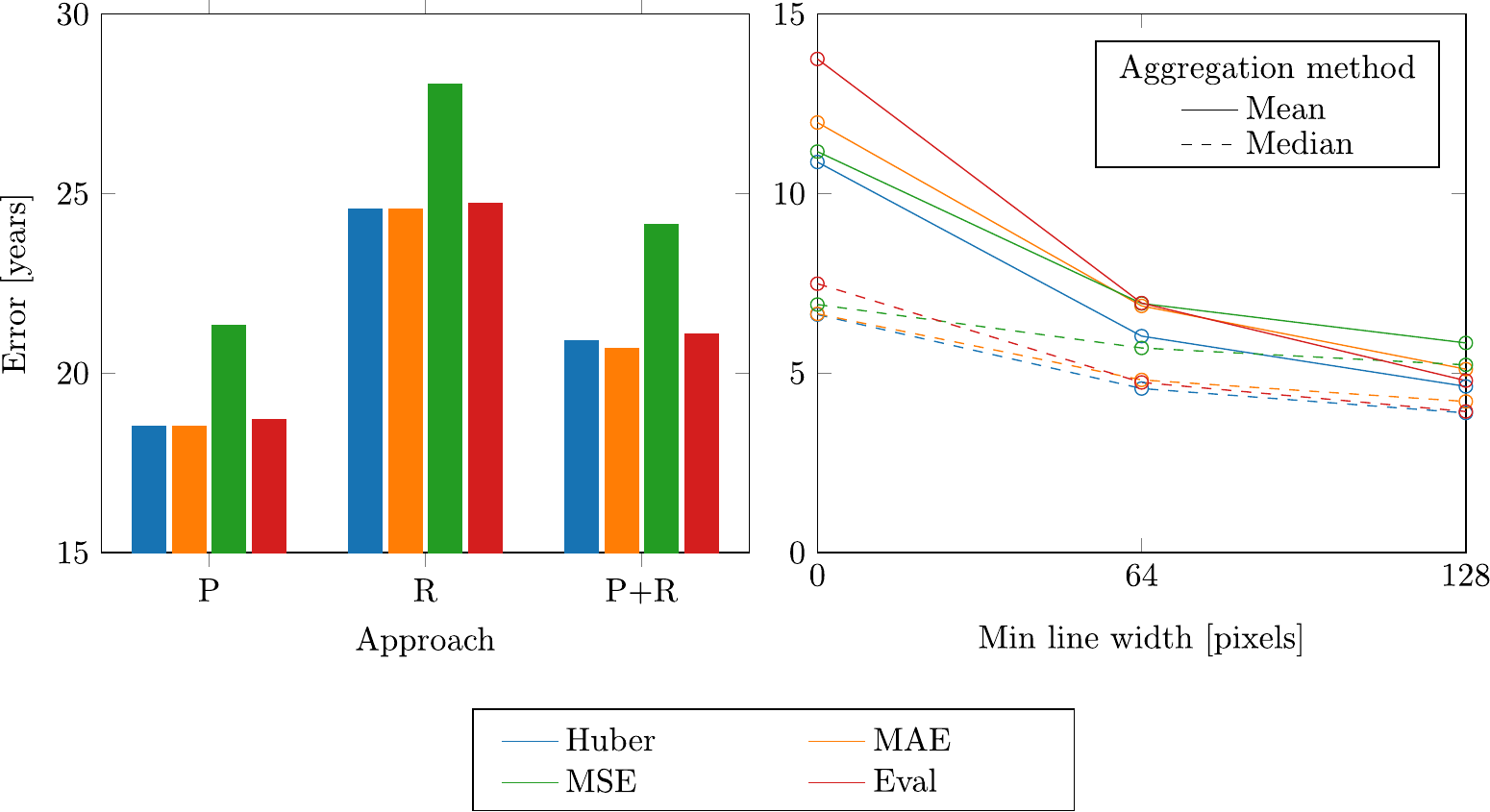}
    \caption{
        Comparison of different date regression loss functions, approaches for patch system (left), and aggregation methods for textline system (right).
        The P, R, P+R stand for whole page approach, text region approach, and whole page + text regions approach, respectively.
        We report the mean absolute error [years] on the validation data.
        Note the different scales on the vertical axes.
    }
    \label{fig:date-aggregations}
\end{figure}

For the dating task, we trained models with each proposed date regression loss function and we evaluated them with all the proposed approaches and aggregation methods.
Results are depicted in Figure~\ref{fig:date-aggregations}.
From these results we decided to use the models trained with the interval Huber loss function.
For the patch system we decided to use the whole pages approach (P) and for the textline system we decided to use the median aggregation method for textlines longer than 128 pixels.

\subsubsection{Classification tasks} 
For the script and font tasks, we trained both systems using both proposed weakly supervised loss functions.
Results of these systems are summarized in Table~\ref{tab:script-font-losses}.
The evaluation was performed on validation sets using the P+R approach for the patch system and mean aggregation method of textlines longer than 64 pixels for the textline system.
We used this configurations based on results of preliminary experiments.

The results show that for the patch system the $L_{soft}$ loss function outperforms the $L_{hard}$ loss function by nearly 50 \% for the font task, while the difference is negligible for the script task.
For the textline system the differences for both tasks are even more negligible.
From these results we decided to use the patch system trained using the $L_{soft}$ loss function and textline system trained using $L_{hard}$ loss function.

\begin{figure}[t]
    \centering
    \includegraphics[width=1\textwidth]{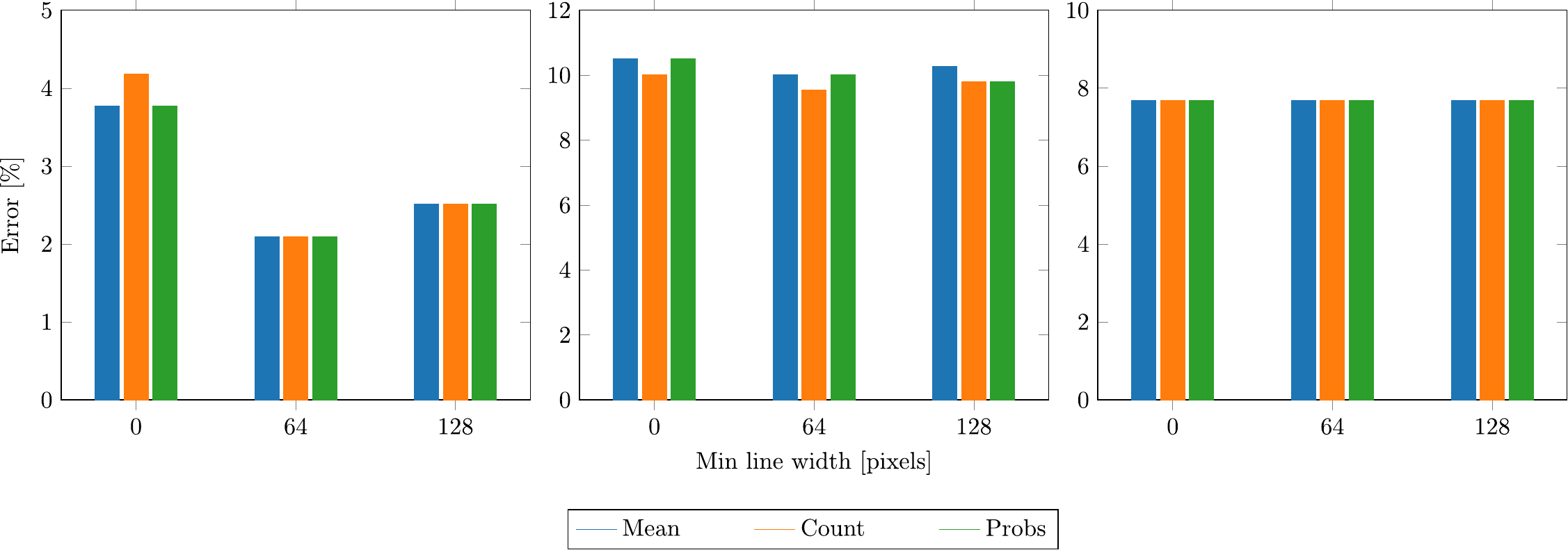}
    \caption{
        Comparison of different aggregation methods for textline system on font (left), script (middle), and localization (right) task.
        We report the error rate on the validation data.
        Note the different scales on the vertical axes.
    }
    \label{fig:line-aggregations}
\end{figure}

For all classification tasks, we also evaluated all approaches described in Section~\ref{sec:patch-system} and all aggregation methods described in Section~\ref{sec:textline-system} on the validation set.
Results of the patch system and textline system are shown in Table~\ref{tab:patch-aggregations} and Figure~\ref{fig:line-aggregations}, respectively.
Based on the results for the patch system, we decided to use the P+R approach for the font and the script classification tasks and the P approach for the localization task.
For the textline system, we selected the mean aggregation method of textlines longer than 64 pixels.

\begin{table}[t]
    \centering
    \caption{%
        Comparison of classification loss functions for training models on dataset with multiple annotations for a single training example.
    }\label{tab:script-font-losses}
    \begin{tabular}{lccccc}
        \toprule
            \multirow{2}{*}{Loss function} & \multicolumn{2}{c}{Patch system} & & \multicolumn{2}{c}{Textline system} \\ 
            \addlinespace[0.05cm] \cline{2-3} \cline{5-6} \addlinespace[0.1cm]
            & \ Font & Script & & \  Font & Script \\
        \midrule
            ${L}_{\mathit{hard}}$     & \  6.22 & 19.09 && \ 2.93 & 10.98 \\
            ${L}_{\mathit{soft}}$     & \  3.32 & 20.29 && \ 2.93 & 10.74 \\
        \bottomrule
    \end{tabular}
\end{table}

\begin{table}[t]
    \centering
    \caption{%
        Comparison of different patch system approaches on classification tasks. 
        The P, R, P+R stand for whole page approach, text region approach, and whole page + text regions approach, respectively.
        We report the error rate on the validation data.
    }\label{tab:patch-aggregations}
    \begin{tabular}{lrrr}
        \toprule
                    \multirow{2}{*}{Task} &
                    \multicolumn{3}{c}{Approach} \\
                    \addlinespace[0.1cm] \cline{2-4} \addlinespace[0.1cm]
                    & \multicolumn{1}{c}{P} & \multicolumn{1}{c}{R} & \multicolumn{1}{c}{P+R}  \\
        \midrule
            Font     &  5.02 &  7.11 & \textbf{ 2.93} \\
            Script   & 17.90 & 25.06 & \textbf{16.47} \\
            Location & 20.00 & 20.00 & \textbf{18.46} \\
        \bottomrule
    \end{tabular}
\end{table}

\subsubsection{System fusion}
We optimized and evaluated both types of system fusion using a 10-fold cross-validation  on our validation sets.
The results presented in Table~\ref{tab:fusion-results} are the averages obtained over those 10 splits.

\begin{table}[t]
    \centering
    \caption{
        Average system fusion results. The values are the average of the results obtained from all possible combinations of validation set splits in the cross-validation.
    }\label{tab:fusion-results}
    
    \begin{tabular}{lcccc}
        \toprule
                                & Font  & Script  & Location  & Date  \\
        \midrule
        Linear fusion           & 1.67  & 9.55    & 7.69      & 4.29 \\
        Log-linear fusion       & 2.09  & 9.55    & 7.69      & ---  \\
        \bottomrule
    \end{tabular}
\end{table}

\subsubsection{Final submission} Finally, our systems were evaluated on the test data by the organizers of the challenge, the results are shown in Table~\ref{tab:test-results}.
Overall, we see that the line system performs much better in classification of script and font type and in dating.
This makes sense as we expect that all the necessary information is within the text content, so allowing the model to work with pre-segmented textlines allows to focus on extracting discriminative features.
On the other hand, the patch system did a better job on localization of documents.
We hypothesise that there are some important visual cues outside of the text in this case, such as ornaments, illustrations, or visual separators.
For the textline system, documents containing little or no text (e.g. title pages) can also be problematic.

When compared to the other systems participating in the ICDAR~2021 Competition on Historical Document Classification, our system achieved the best results in each task and became the overall winner of the competition.

\subsubsection{Checkpoints averaging}
In an attempt to mitigate the effect of spurious changes in validation error coming from random steps of gradient descent, we introduce checkpoint averaging for our models used for the final submission.
For the patch system, we took the best performing checkpoint by validation performance and averaged it with checkpoints $\pm 1,2,3$ thousand updates apart, for a total of 7 model averaged.
For textline system, we averaged the last $N$ checkpoints from the training.
Here, $N$ was tuned to optimize validation performance in range $1 < N < 10$.

\begin{table}[t]
    \centering
    \caption{
        Final results on the ICDAR 2021 Competition on Historical Document Classification test data.
        Font, script and location are reported as accuracy [\%]; for date, mean absolute error is given [years].
        All of the presented results have been measured by the challenge organizers.
        The first four lines represent our submissions. 
        The results of the other teams that participated in the competition~\cite{seuret_icdar_2021} are shown below the line. 
        Since each team could have participated with multiple systems, we present here only the best result for each task for these teams.
    }\label{tab:test-results}
    
    \begin{tabular}{lcccc}
        \toprule
                                & Font              & Script            & Location          & Date              \\
        \midrule
        Textline system         & 98.42             & 88.54             & 69.85             & \textbf{21.91}    \\
        Patch system            & 95.68             & 80.26             & 75.08             & 32.45             \\[2pt]
        Linear fusion           & 98.27             & \textbf{88.84}    & 70.77             & 21.99             \\
        Log-linear fusion       & \textbf{98.48}    & 88.60             & \textbf{79.69}    & ---               \\
        \midrule
        Baseline                & ---               & 55.22             & 62.46             & ---               \\
        The North LTU           & 82.80             & 74.12             & 43.69             & 79.43             \\
        CLUZH                   & 95.66             & 35.25             & ---               & ---               \\
        NAVER Papago            & 97.17             & ---               & ---               & ---               \\
        \bottomrule
    \end{tabular}
\end{table}

\section{Conclusion}

We proposed a system for visual document classification fusing patch level and textline level approaches.
The experiments on the datasets from ICDAR 2021 Competition on Historical Document Classification show that line level processing significantly outperforms patch processing in classification of font and script types and in dating. 
On the other hand, the patch level approach yields better localization results.

Our proposed weakly supervised classification loss functions proved effective in utilizing multiple page-level classification labels. 
Similarly for the dating problem, where the labels are given as intervals, specific interval loss functions brought improvement over baseline regression approaches; namely the proposed interval Huber loss yields best results.
The fusion provided mediocre improvements.
It improved results significantly only in the localization task, the other tasks were dominated by the textline system.
The small improvement due to fusion indicates that the patch and line level systems have similar strengths and weaknesses. 

We believe the the proposed approach could be further improved by combining the task-specific networks into a single multi-task network. 
Further, text surely contains important cues for dating and localization and we intend to add a text-level system working with automatic text transcriptions, and possibly utilizing exiting historical text corpora.

\subsubsection{Acknowledgement}
This work has been supported by the Ministry of Culture Czech Republic in NAKI II project PERO (DG18P02OVV055) and by Czech National Science Foundation (GACR) project ”NEUREM3” No. 19-26934X.

\bibliographystyle{splncs04}
\bibliography{bibliography}

\end{document}